\documentclass[10pt, a4paper]{article}
\usepackage{lrec}
\usepackage{multibib}
\newcites{languageresource}{Language Resources}
\usepackage{graphicx}
\usepackage{tabularx}
\usepackage{soul}

\usepackage{epstopdf}
\usepackage[utf8]{inputenc}

\usepackage{hyperref}
\usepackage{xstring}

\usepackage{color}

\title{SOLO: A Corpus of Tweets for Examining the State of Being Alone}

\name{Svetlana Kiritchenko$^1$, Will E. Hipson$^2$, Robert J. Coplan$^2$, Saif M. Mohammad$^1$}

\address{$^1$National Research Council Canada, $^2$Carleton University, Ottawa, Canada\\
         svetlana.kiritchenko@nrc-cnrc.gc.ca, williamhipson@cmail.carleton.ca,\\ robert.coplan@carleton.ca, saif.mohammad@nrc-cnrc.gc.ca\\}

\abstract{The state of being alone can have a substantial impact on our lives, though experiences with time alone diverge significantly among individuals.  
Psychologists distinguish between the concept of \textit{solitude}, a positive state of voluntary aloneness, and the concept of \textit{loneliness}, a negative state of dissatisfaction with the quality of one’s social interactions. 
Here, for the first time, we conduct a large-scale computational analysis to explore how the terms associated with the state of being alone are used in online language. 
We present SOLO (\textit{\underline{S}tate \underline{o}f Being A\underline{lo}ne}), a corpus of over 4 million tweets collected with query terms \textit{solitude}, \textit{lonely}, and \textit{loneliness}. 
We use SOLO to analyze the language and emotions associated with the state of being alone. 
We show that the term \textit{solitude} tends to co-occur with more positive, high-dominance words (e.g., \textit{enjoy}, \textit{bliss}) while the terms \textit{lonely} and \textit{loneliness} frequently co-occur with negative, low-dominance words (e.g., \textit{scared}, \textit{depressed}), which confirms the conceptual distinctions made in psychology. 
We also show that women are more likely to report on negative feelings of being lonely as compared to men, and there are more teenagers among the tweeters that use the word \textit{lonely} than among the tweeters that use the word \textit{solitude}.
\\ 
\newline \Keywords{solitude, lonely, mental health, well-being} }

\begin{document}

\maketitleabstract

\newcommand{\assocscore}[1]{{\rm \it Score}\,(#1)}
\newcommand{\pmiscore}[2]{{\rm \it PMI}\,({\rm \it #1}, {\rm \it #2})}
\newcommand{\freqi}[1]{{\rm \it freq}\,({\rm \it #1})}
\newcommand{\freqij}[2]{{\rm \it freq}\,({\rm \it #1}, {\rm \it #2})}

\section{Introduction}
 We have all experienced the state of being alone one time or another:  
perhaps, a loved one was away, or
our Instagram post did not stir up a barrage of likes, or 
we enjoyed a quiet hike, or we felt disconnected from those around us.
Further, older people and young adults experience loneliness at markedly higher rates
than others \cite{luhmann2016age,hawkley2015perceived}.

The state of being alone can have a substantial impact on our lives. On the one hand, \textit{loneliness}---a negative and unwanted state of being alone---has been shown to be correlated with increased cognitive decline, dementia, depression, suicide ideation, self harm, and even death \cite{gerst2015loneliness,hawkley2015perceived,luo2012loneliness,endo2017preference}.\footnote{The negative public health impacts of loneliness are so great that in 2018 the UK appointed a minister for loneliness.}
On the other hand, \textit{solitude}---a positive and self-driven state of being alone---has been shown to improve autonomy, creativity, and well-being \cite{long2003,knafo2012alone,coplan2017should,coplan2019seeking}.
Loneliness and solitude have also been shown to play a role in the adaptive fitness of our species \cite{hawkley2015perceived,larson1990solitary}.
Thus loneliness and solitude are starting to receive substantial amounts of attention from the medical and psychological research. 
Yet, there is no large-scale computational work on analyzing the language of being alone. 

Here, for the first time, we present a large corpus of tweets associated with the state of being alone.
We will refer to it as the \textit{\underline{S}tate \underline{o}f Being A\underline{lo}ne} corpus, or \textit{SOLO} for short. SOLO includes over 4 million tweets, each of which consists of at least one of the following tokens: \textit{solitude}, \textit{lonely}, and \textit{loneliness}. We use SOLO to analyze the language and emotions associated with the state of being alone. Specifically, we explore the following  questions:\\ 
\begin{itemize}
    \item When people use terms such as {\it solitude, alone}, and {\it loneliness} in tweets, how often are they referring to the state of being alone as opposed to some other sense of those words?\\[-18pt]
    \item Do we find evidence from the text that solitude is indeed more self-driven than loneliness (as theorized by psychologists)?\\[-18pt]
    \item Do we find evidence from the text that the speakers view solitude as a more positive concept than loneliness (as theorized by psychologists)?\\[-18pt]
    \item Which words are associated with solitude, and which words are associated with loneliness?\\[-18pt] 
    \item Do different demographic groups (e.g., different genders, age groups, etc.) perceive solitude and loneliness differently? \\[-16pt]
\end{itemize}
\noindent Most of the past studies exploring such questions come from Psychology (see next section). They involve self-reports from a small number of people. Here, for the first time, we computationally examine millions of tweets associated with the state of being alone for the language used, and especially the emotion associations.
We also make SOLO freely available for research.\footnote{https://svkir.com/projects/solo.html} 
We hope that this new dataset will bring fresh  attention to the relationship between the state of being alone and our well-being.

\section{Related Work}
\label{related-work}

Time spent alone can have varying emotional effects. For instance, time alone 
is experienced negatively in those cases when we are unable to fulfill our needs for social interaction \cite{baumeister1995need}, but positively when we are exhausted from long periods of social interaction and desire time for relaxation and reflection \cite{nguyen2018solitude,long2003}. Given that an estimated 25--33\% of waking time is spent being alone\\ 
\cite{larson1982time}, identifying and distinguishing between `positive' and `negative' instances of being alone has substantial implications for improving people's well-being.

Many theoretical perspectives have emerged to explain these divergent experiences of being alone. Proponents of self-determination theory \cite{deci2010self} postulate that time alone that is intrinsically motivated (i.e., \textit{choosing} to spend time alone) is better for one’s well-being than time alone that arises for external reasons (e.g., one who is alone due to the nature of their work) \cite{chua2008self,nguyen2018solitude}. 

However, the experience of being alone may also differ as a result of when this state arises. Someone who spends a lot of time alone may come to feel lonely because they perceive their social network as deficient \cite{hawkley2010loneliness}, in which case subsequent moments in solitude are likely to diminish in pleasantness. Conversely, someone who is inundated with social activity may become dissatisfied with the amount of time they get to spend alone \cite{coplan2019seeking}, in which case being alone would be experienced as even more pleasant than usual. 
As far as we know, there has been no large-scale computational work examining text associated with the state of being alone.

Even though emotions are central to human experience and they have been studied for centuries, there are still many unknowns about their inner workings. Two prominent models of emotions are the dimensional  model and the basic emotions model.  
As per the dimensional model \cite{Osgood1957,russell1980circumplex,russell2003core},  emotions are points in a three-dimensional space of valence (positive--negative), arousal (active--passive), and dominance (dominant--submissive). 
Thus, when comparing the meanings of two words, we can compare their degrees of valence, arousal, or dominance. 
For example, the word {\it party} indicates more positiveness  than the word {\it crying}; {\it terrible} indicates  more arousal than {\it conversation}; and {\it hike} indicates more dominance than  {\it abandoned}.

According to the basic emotions model (aka discrete model) \cite{Ekman92,Plutchik80,frijda1988laws}, some emotions, such as joy, sadness, fear, etc., are more basic than others, and these emotions are each to be treated as separate categories. 

We use the NRC Valence, Arousal, and Dominance (NRC VAD) lexicon \cite{vad-acl2018} and the NRC Emotion lexicon \cite{Mohammad13,mohammad2010} to determine the emotion associations of the words in SOLO. These lexicons were created by manual annotation. The NRC VAD lexicon has valence, arousal, and dominance scores for over twenty thousand English terms, and it was created using a comparative annotation technique called Best-Worst Scaling (BWS) \cite{Louviere_1991,Louviere2015,maxdiff-naacl2016}.
It has been shown to have high reliability (repeated annotations produce similar association scores).
The NRC Emotion lexicon has binary (associated or not associated) scores for about fourteen thousand English terms (a subset of terms in the VAD lexicon) with eight basic emotions (joy, sadness, fear, anger, surprise, anticipation, disgust, and trust) as well as positive and negative sentiment.

\section{Creating the SOLO Corpus}

We now describe how we collected tweets related to the state of being alone and created the SOLO corpus. 

\subsection{Query Term Selection}

After consulting with psychologists on our team and utilizing different thesauri, we created a list of words and short phrases related to the state of being alone: \textit{alone}, \textit{alone time}, \textit{aloneness}, \textit{confinement}, \textit{desert}, \textit{detachment}, \textit{get away from it all}, \textit{get away from people}, \textit{hermit}, \textit{isolation}, \textit{loneliness}, \textit{lonely}, \textit{lonesomeness}, \textit{me time}, \textit{peace and quiet}, \textit{privacy}, \textit{quarantine}, \textit{reclusiveness}, \textit{retirement}, \textit{seclusion}, \textit{separateness}, \textit{serenity}, \textit{silence}, \textit{solitariness}, \textit{solitude}, \textit{tranquility}, \textit{undisturbed}, \textit{wilderness}, \textit{withdrawal}. 
We collected tweets using these query terms for a few weeks, and then manually checked the relevance of the obtained tweets. 
Some query terms (e.g., \textit{solitariness}, \textit{reclusiveness}, \textit{lonesomeness}, \textit{aloneness}, \textit{get away from it all}) were rarely used on Twitter and, therefore, were discarded. 
Some terms (e.g., \textit{silence}, \textit{privacy}, \textit{retirement}, \textit{desert}) were often used in other senses, not related to the state of being alone. 
Even for the query word \textit{alone}, only about half of the collected tweets related to the concept of being alone. 
In many tweets, \textit{alone} was used for emphasis (e.g., ``\textit{only you and you alone can thrill me like you do}'', ``\textit{I barely like Christmas music on Christmas lol, let alone in early November}''). 
After this manual inspection, we decided to keep three terms: \textit{solitude} and \textit{loneliness} (nouns), and  \textit{lonely} (adjective).

\subsection{Collecting Tweets}

\vspace{3pt}
\noindent \textbf{SOLO Corpus:} Tweets related to the state of being alone were collected by polling the Twitter API from August 28, 2018 to July 10, 2019 with the following query terms: \textit{loneliness}, \textit{lonely}, and \textit{solitude}. 
We discarded duplicate tweets, short tweets (containing less than three words), and tweets with external URLs. 
Further, we kept only up to three tweets per user. 
This minimizes the impact of prolific tweeters and bots on the corpus. 
We refer to the combined set of the remaining tweets as the \textit{\underline{S}tate \underline{o}f being A\underline{lo}ne} corpus, or \textit{SOLO} for short. 
We refer to the individual sets of tweets as the \textit{loneliness sub-corpus}, the \textit{lonely sub-corpus}, and the \textit{solitude sub-corpus}, respectively. 
Table~\ref{tab:num-tweets} shows the number of tweets in each sub-corpus. 
In total, the SOLO Corpus contains over four million tweets.

\vspace{5pt}
\noindent \textbf{General Tweets:} As a control corpus, we collected tweets by polling the Twitter API from May 16, 2019 until June 12, 2019 using English function words (e.g., \textit{is}, \textit{on}, \textit{they}, etc.) as query terms. 
Again, we discarded duplicate tweets, short tweets (containing less than three words), tweets with external URLs, and kept only up to three tweets per user.
We will refer to this set of tweets as the \textit{General Tweet Corpus}. 
It includes over 21 million tweets. 

\begin{table}[t]
{\small
\begin{center}
\begin{tabular}{lrr}
\hline
\textbf{Corpus} & \textbf{\# of tweets} & \textbf{\# of users}\\
\hline
SOLO Corpus:\\
$\ \ $ loneliness  &    489,264 & 408,659\\
$\ \ $ lonely & 3,339,166 & 2,443,210\\
$\ \ $ solitude     &   191,643 & 158,878\\ [5pt]
$\ \ $ All & 4,020,073 & 3,010,747\\ [5pt]
General Tweet Corpus & 21,719,409 & 12,096,240\\[3pt]
      \hline
\textbf{Total} & \textbf{25,739,482} & \textbf{15,106,987}\\
      \hline
\end{tabular}
\caption{The number of tweets for each query term.} 
\label{tab:num-tweets}
 \end{center}
 }
\end{table}

\subsection{Tweet Volume}

For the same time period (about a year), we were able to collect seventeen times more tweets with the word \textit{lonely} and two-and-a-half times more tweets with the word \textit{loneliness} than tweets 
with the word \textit{solitude}. 
This suggests that most users refer to the state of being alone through the use of words \textit{lonely} and \textit{loneliness}, and rarely 
using the word \textit{solitude}.  
In a period of one year, close to three million users posted at least one tweet with the words \textit{lonely} or \textit{loneliness}, which reflects the magnitude of the loneliness problem.

\section{Assessing Relevance of the SOLO Tweets to the State of Being Alone}
\label{tweet_relevancy}

\begin{table}[t]
{\small
\begin{center}
\begin{tabular}{lr}
\hline
\textbf{Corpus} & \textbf{Percentage of relevant tweets}\\
\hline
loneliness  &   93\% \\
lonely & 96\%\\
solitude     &   92\%\\ [5pt]
Average & 94\%\\ 
      \hline
\end{tabular}
\caption{Percentage of relevant tweets for each query term.}
\label{tab:relevant-tweets}
 \end{center}
 }
 \vspace*{-3mm}
\end{table}

A tweet may include the term {\it loneliness}, {\it lonely}, or \textit{solitude} and yet may not be relevant to the state of being alone. Thus we manually examined a small sample of SOLO to determine the percentage of relevant tweets.
We considered a tweet to be relevant if it directly referred to the state of being alone. 
This included (but was not limited to):\\[-12pt]
\begin{itemize}
	\item a personal statement about being alone,\\ [-12pt]
	\item a statement about other people being alone,\\ [-12pt]
	\item a general statement about aspects of being alone,\\[-12pt]
	\item a message of support (e.g., ``\textit{you are not alone}''),\\[-12pt]
	\item a quote from literature about being alone.\\[-10pt]
\end{itemize}
We considered tweets to be irrelevant if the query word (\textit{loneliness}, \textit{lonely}, \textit{solitude}) was used as part of a title (of a book, song, etc.) or a name (of a place, a stadium, etc.). 
Tweets containing advertisements were also considered irrelevant. 

For each query term, we randomly selected 100 tweets with that term and counted the percentage of relevant tweets. 
Table~\ref{tab:relevant-tweets} shows the results. 
Observe that for all the query terms, over 90\% of examined tweets were relevant to the state of being alone. 
This confirms the suitability of the SOLO Corpus for studying the everyday language associated with the state of being alone.

\section{Analyzing the Language and Emotions Associated with the State of Being Alone}

We examine the language of the SOLO tweets to determine if the concept words \textit{loneliness}, \textit{lonely}, and \textit{solitude} tend to be used in different emotional contexts. 
In particular, we explore the question of whether Twitter users perceive the concept of solitude as more positive and self-driven and the concept of loneliness as more negative and externally imposed as suggested by psychology literature.  
For this, in Section 5.1, we manually analyze a sample of tweets for the types of contexts in which people use the words \textit{loneliness}, \textit{lonely}, and \textit{solitude}.
We also computationally identify and compare words strongly associated with each of these terms. 
In Section 5.2, we examine the words occurring in SOLO for their emotional associations.

\subsection{Language Associated with Being Alone}
\label{lang-analysis}

\begin{table}[t]
{\small
\begin{center}
\begin{tabular}{lrrr}
\hline
\textbf{Categories} & \textbf{loneliness} & \textbf{lonely} & \textbf{solitude}\\

\hline
first-hand experience & 0.35 & 0.62 & 0.47\\
other people's experience & 0.15 & 0.16 & 0.09\\
general statement & 0.30 & 0.09 & 0.21\\
literary quote & 0.19 & 0.06 & 0.16\\
offering support & 0.00 & 0.01 & 0.05\\
other & 0.01 & 0.06 & 0.02\\
      \hline
\end{tabular}
\caption{Different types of SOLO tweets and their relative frequency in each sub-corpus.}
\label{tab:themes}
 \end{center}
 }
\end{table}

\setlength{\tabcolsep}{3pt}

\begin{table*}[t]
{\small
\begin{center}
\begin{tabular}{ll}
\hline
\textbf{SOLO term} &\textbf{Words associated with the term}\\
\hline
loneliness & \textit{alone, feeling, lonely, depression, pain, sadness, isolation, fear, killing, feelings, anxiety, happiness, cure,} \\
&          \textit{solitude, hurts, emptiness, crippling, anger, silence, fill, suffering, relationships, empty, darkness, boredom}\\ [5pt]
lonely &  \textit{feel, sad, feeling, alone, friends, sometimes, single, felt, bored, feels, nights, scared, depressed, af, cold,} \\
&          \textit{island, christmas, empty, hearts, loneliness, miserable, surrounded, horny, asf, desperate}\\ [5pt]
solitude & \textit{alone, enjoy, peace, silence, loneliness, fortress, quiet, hundred, lonely, enjoying, comfort, prefer, nature,} \\
 &         \textit{isolation, comfortable, bliss, moments, sea, presence, peaceful, seek, embrace, darkness, gabriel, inner}\\
      \hline
\end{tabular}
\caption{The most frequent words strongly associated with the terms \textit{loneliness}, \textit{lonely}, and \textit{solitude}.}
\label{tab:lex-words}
 \end{center}
 }
\end{table*}

First, we look at how people use the terms \textit{loneliness}, \textit{lonely}, and \textit{solitude} in everyday language of tweets. Do people often describe their own feelings and experiences or offer support to other people? Do they just make general statements about different aspects of being alone? Which words are most likely to co-occur with these terms?

\vspace{5pt}
\noindent \textbf{Manual Examination of the SOLO Tweets:} 
We manually examined randomly selected samples of 100 tweets from the loneliness, lonely, and solitude sub-corpora to identify the types of messages users are likely to post using these terms. 
Table~\ref{tab:themes} shows the results. 

In tweets with the word \textit{solitude}, people often describe their own experiences and attitudes (e.g., ``\textit{I fell in love with my solitude.. everything changed after that.}''), provide general statements about positive or negative aspects of being alone (e.g., ``\textit{Solitude can be either comforting or really painful.}''), and cite relevant quotes from notable people and literary sources (e.g., ``\textit{The monotony and solitude of a quiet life stimulates the creative mind - Albert Einstein}''). 
They less often discuss other people's experiences (e.g., ``\textit{It seems like they hate everything that isn't profitable - whether it's wolves, wild horses, stunning landscapes, solitude...}'') or offer support (e.g., ``\textit{that is saaaaaddd, but don't worry, solitude is a nice friend}'').  

When people use the word \textit{lonely}, they mostly report on their own feelings (e.g., ``\textit{Feeling lonely and forgotten :/}'') and those of other people (e.g., ``\textit{Well you were clearly very lonely.}''). 
In tweets with the word \textit{loneliness}, users less often describe their own experiences (e.g., ``\textit{The level of loneliness I've reached is at an all time high}''), and more often make general statements (e.g., ``\textit{we don't know how to appreciate loneliness}'') and quote celebrities and literary sources (e.g., ``\textit{If you are afraid of loneliness, don't marry. -Anton Chekhov}''), than in tweets with the words \textit{lonely} and \textit{solitude}. 

Notably, in 14\% of tweets from the solitude sample, tweeters explicitly assert their need to spend some time alone to reflect, heal, or focus on important tasks (e.g., ``\textit{It's funny how the universe works...this moment of solitude was unplanned but definitely needed.}'').

\vspace{10pt}
\noindent \textbf{Words Associated with Loneliness, Lonely, and Solitude:} We identify words that are associated with the SOLO query terms, \textit{loneliness}, \textit{lonely}, and \textit{solitude}, i.e., words that tend to appear in tweets with these query terms more often than they do in the General Tweet Corpus. 
For this, we calculate an association score of a word $w$ with the target sub-corpus $C_{target}$ ($ {\rm \it target} \in \left\{ {\rm \it loneliness, lonely, solitude} \right\}$) as compared to the corpus of general tweets (the reference corpus, $C_{reference}$):
\vspace{3pt}
\begin{equation}
\label{eq-score1}
\assocscore{w} = \pmiscore{w}{C_{target}} - \pmiscore{w}{C_{reference}}
\end{equation}

\vspace{3pt}
\noindent PMI stands for pointwise mutual information:
\vspace{3pt}
\begin{equation}
\label{eq-score2}
\pmiscore{w}{C_{target}} = log_{2} \ \frac{\freqij{w}{C_{target}} * N}{\freqi{w} * \freqi{C_{target}}}
\end{equation}

\vspace{3pt}
\noindent where \freqij{w}{$C_{target}$} is the number of times the word $w$ occurs in the target corpus, \freqi{w} is the total frequency of the word $w$ in the two corpora (target and reference), \freqi{$C_{target}$} is the total number of words in the target corpus, and $N$ is the total number of words in the two corpora.
\pmiscore{w}{$C_{reference}$} is calculated in a similar way.
Thus, Equation~\ref{eq-score1} is simplified to:
\vspace{3pt}
\begin{equation}
\label{eq-score3}
\assocscore{w} = log_{2} \ \frac{\freqij{w}{C_{target}} * \freqi{C_{reference}}}{\freqij{w}{C_{reference}} * \freqi{C_{target}}}
\end{equation}

\vspace{3pt}
Since PMI is known to be a poor estimator of association for low-frequency events, we ignore terms that occur less than 25 times in total in both corpora. 

Association scores can range from $-\infty$ to $+\infty$; in practice, however, they usually range from around $-6$ to $6$. 
A positive score indicates a greater overall association with the target corpus, that is the word appears at a higher rate (more occurrences per 100 words) in the target corpus than in the reference corpus. 
A negative score indicates that a word appears at a lower rate in the target corpus than in the reference corpus. 
The magnitude is indicative of the degree of association.
Note that there exist numerous other methods to estimate the degree of association of a word with a category (e.g., cross entropy, Chi-squared test, and information gain).
We have chosen PMI because it is simple and robust and has been successfully applied in a number of NLP tasks \cite{clark2016combining,kiritchenko2014sentiment}.

We calculate association scores with the loneliness, lonely, and solitude sub-corpora for all words in the SOLO corpus. 
We say that a word is \textbf{strongly associated} with a sub-corpus if the corresponding association score is greater than or equal to $1.5$.\footnote{The threshold of 1.5 is somewhat arbitrary, but reasonable.} 
Table~\ref{tab:lex-words} shows 25 most frequent words in the loneliness, lonely, and solitude sub-corpora that are strongly associated with them. 
Observe that the words strongly associated with \textit{solitude} are mostly positive. 
Tweets in the solitude sub-corpus tend to describe peaceful, enjoyable moments, often in the natural surroundings. 
The presence of high-dominance words, such as \textit{enjoy}, \textit{prefer}, and \textit{comfort}, indicate that the person most likely feels in control over a situation, that the time alone was self imposed and desirable. 
Words strongly associated with \textit{lonely} and \textit{loneliness}, on the other hand, are mostly negative and low in dominance. 
These tweets often refer to the feelings of sadness, anxiety, depression, and boredom. 
Words like \textit{friends}, \textit{relationships}, and \textit{Christmas} probably reflect the unfulfilled need for social interaction that is often felt more strongly during traditional family holidays like Christmas. 

\begin{table*}[t]
{\small
\begin{center}
\begin{tabular}{ll}
\hline
\textbf{SOLO term} & \textbf{Words associated with the term}\\
\hline
solitude & \textit{enjoy, peace, silence, fortress, quiet, hundred, enjoying, prefer, nature, bliss, complete, presence, peaceful,}\\
&  \textit{seek, embrace, gabriel, inner, marquez, value, spiritual, noise, superman, competing, recharge, prayer}\\ [5pt]
lonely \& &   \textit{feeling, im, sad, girl, ass, nights, lmao, bitch, boy, baby, scared, bored, girls, hi, cuz, somebody, depressed,}\\
loneliness & \textit{hearts, sucks, broke, club, af, pls, hurts, cute}\\
      \hline
\end{tabular}
\caption{The most frequent words strongly associated with \textit{solitude} as opposed to \textit{lonely} and \textit{loneliness}.}
\label{tab:lex-opp-words}
 \end{center}
 }
\end{table*}

\setlength{\tabcolsep}{6pt}

\vspace{10pt}
\noindent \textbf{Solitude--Loneliness Dimension of Word Association:} 
We can use the \textit{solitude} corpus to study how people talk about solitude.
Similarly, we can use the \textit{lonely} and \textit{loneliness} corpora (jointly) to study how people talk about loneliness.\footnote{We use the italicized term (e.g., \textit{loneliness}) to refer to the query term, and the non-italicized form (e.g., loneliness) to refer to the mental/physical state.}
In the sub-section above, we explored each of the query term sub-corpora in comparison with the General Tweets Corpus. Here, in order to determine the extent to which words are associated with solitude as opposed to loneliness, we calculate the solitude--loneliness association score as shown below:\\
\begin{equation}
\label{eq-score4}
\assocscore{w} = \pmiscore{w}{C_{solitude}} - \pmiscore{w}{C_{loneliness}}
\end{equation}

Using this score we can place words along the solitude--loneliness dimension,
where words strongly associated with solitude but not with loneliness are towards one end and 
words strongly associated with loneliness but not with solitude are towards the other end. 

Table~\ref{tab:lex-opp-words} shows 25 most frequent words that are more strongly associated with solitude than with loneliness (solitude--loneliness association score $\geq 1.5$), and 25 most frequent words that are more strongly associated with loneliness than with solitude (solitude--loneliness association score $\leq -1.5$). 
Observe that words that are more strongly associated with solitude than with loneliness are positive and high dominance words. 
These are words referring to peaceful and spiritual activities of being with oneself, recharging, and enjoying the present moment. 
In contrast, the words more strongly associated with loneliness than with solitude refer to negative personal experiences of being sad, scared, bored, hurt, and broken-hearted. 

\vspace{5pt} 
\subsection{Emotions Associated with Being Alone}
\label{emotion-analysis}

\begin{table}[t]
{\small
\begin{center}
\begin{tabular}{lrrr}
\hline
\textbf{Sentiment} & \textbf{loneliness} & \textbf{lonely} & \textbf{solitude}\\
\hline
positive & 0.05 & 0.03 & 0.71\\
negative & 0.71 & 0.84 & 0.11\\
mixed & 0.14 & 0.06 & 0.18\\ 
unclear & 0.10 & 0.07 & 0\\
      \hline
\end{tabular}
\caption{Proportions of the SOLO tweets with different sentiments towards the state of being alone.}
\label{tab:sent}
 \end{center}
 }
\end{table}

In this section, we measure the emotional context in which the SOLO query terms, \textit{loneliness}, \textit{lonely}, and \textit{solitude}, occur. 
In particular, we investigate whether people use these terms in different emotional contexts 
and whether they are associated with the qualities suggested in the psychology literature. 
We analyze a sample of the SOLO corpus manually and the full corpus computationally using existing word--emotion association lexicons. 

\vspace{5pt}
\noindent \textbf{Manual Examination of Sentiment in the SOLO Corpus:} 
We randomly sampled 100 tweets each from the loneliness, lonely, and solitude sub-corpora, and manually examined each of these tweets to determine whether
they express positive, negative, or mixed attitudes towards the state of being alone. 
Table~\ref{tab:sent} shows the results. 

Observe that tweeters that use the term \textit{solitude} mostly have a positive attitude towards being alone (e.g., ``\textit{Have you ever: felt lonely? No, I love my solitude.}''), yet sometimes mixed (e.g., ``\textit{What is the balance for those of us that love the solitude but wanna have companionship ??}'') or even negative (e.g., ``\textit{Some people prefer to live in solitude, but no one can withstand it}'') sentiments can be expressed.  
On the other hand, the vast majority of tweeters that use the words \textit{lonely} and \textit{loneliness} have a negative attitude towards being alone (e.g., ``\textit{i'm really lonely and really sad}''). 
Only rarely do people include the words \textit{lonely} and \textit{loneliness} when they express positive sentiments in the SOLO tweets (e.g., ``\textit{Loneliness is designed to help you discover who you are ... and to stop looking outside yourself for your worth. ? Mandy Hal}'').

\vspace{5pt}
\noindent \textbf{Basic Emotions Associated with Words in SOLO:} Next, we look at the whole SOLO Corpus and analyze emotions associated with words occurring in the SOLO tweets. 
We use the NRC Word--Emotion Association Lexicon \cite{Mohammad13,mohammad2010} which has entries for over 14,000 English common words.\footnote{http://saifmohammad.com/WebPages/NRC-Emotion-Lexicon.htm}
It provides labels for eight basic emotions (anger, fear, sadness, disgust, joy, anticipation, surprise, and trust) and two sentiments (positive and negative).  
The labels are binary indicating whether a word is associated with an emotion (or sentiment) or not. 
The lexicon was created by crowd-sourcing the annotations. 
We consider only those words in SOLO that appear in the lexicon, and count the percentage of words associated with each emotion (i.e., out of every 100 words, how many are associated with sadness, joy, etc.). (The SOLO query words are excluded from the analysis.)  

\begin{figure}[t]
\begin{center}
\includegraphics[scale=0.34]{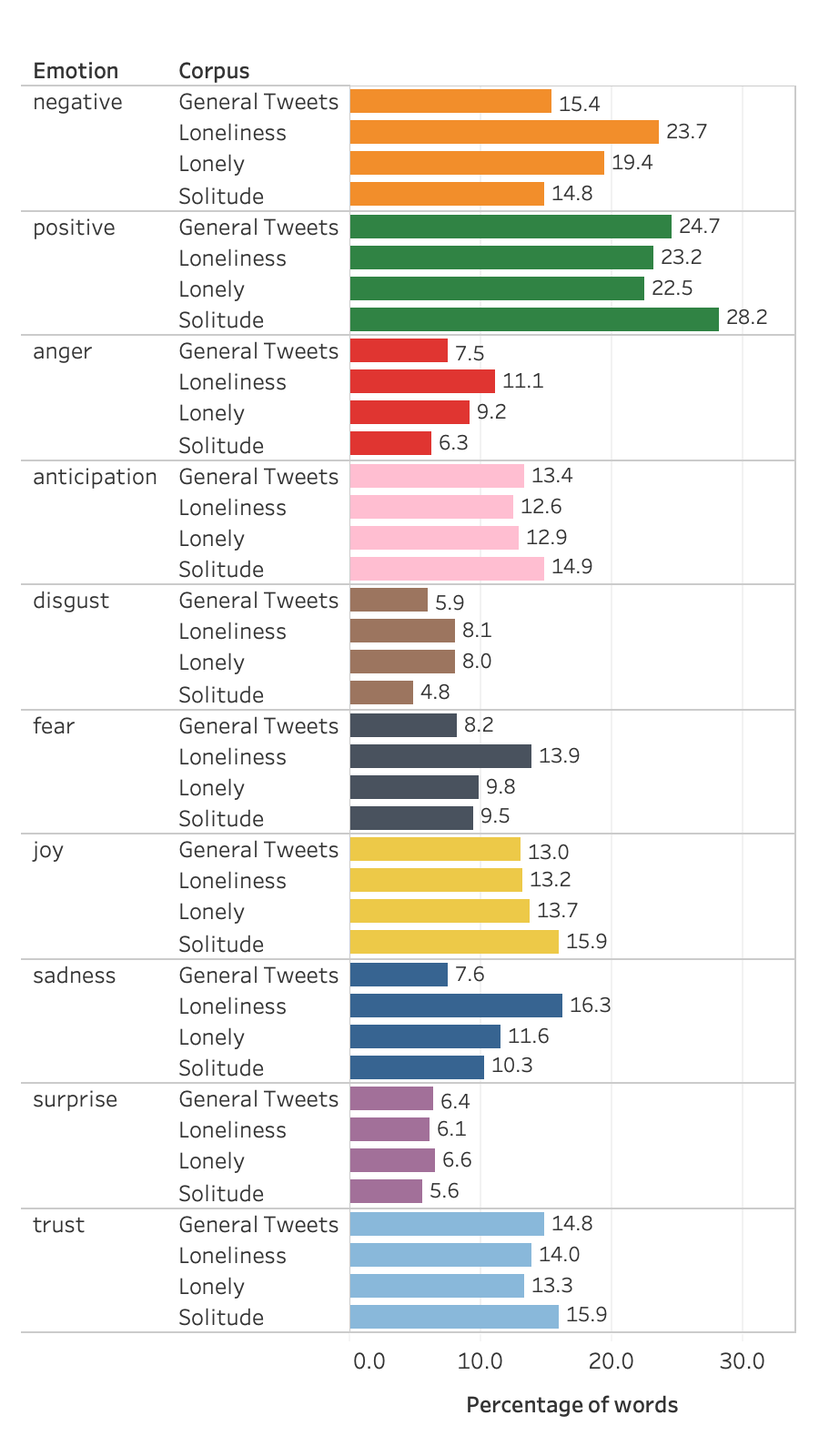} 
\caption{The percentage of words associated with eight basic emotions in different sub-corpora.}
\label{fig-EmoLex}
\end{center}
\end{figure}

Figure~\ref{fig-EmoLex} shows the results for the different sub-corpora of the SOLO corpus. 
For comparison, we also show the results for the General Tweets Corpus. 
For each emotion, the differences between the word percentages for the sub-corpora are statistically significant (Chi-squared test,  $p < 0.0001$).
Observe that tweets in the solitude sub-corpus contain more words associated with the positive sentiment and more words associated with the emotions of joy, anticipation, and trust than the tweets in other sub-corpora, including the general tweets. 
There are 25--30\% more positive words in the solitude tweets than in the lonely and loneliness tweets. 
On the other hand, tweets with the words \textit{lonely} and \textit{loneliness} have more words associated with the negative sentiment and more words associated with the emotions of anger, fear, sadness, and disgust. 
There are 60\% more negative words in the loneliness tweets than in the solitude or the general tweets. 
Somewhat surprisingly, tweets in the loneliness sub-corpus have significantly more (20--40\%) words associated with the negative sentiment and the negative emotions of anger, fear, and sadness than tweets in the lonely sub-corpus. 

\vspace{5pt}
\noindent \textbf{Valence-Arousal-Dominance of Words in SOLO:} To analyze the SOLO corpus with regard to the dimensional theory of emotions, we use the NRC Valence, Arousal, and Dominance (VAD) Lexicon \cite{vad-acl2018}.\footnote{http://saifmohammad.com/WebPages/nrc-vad.html}
The VAD lexicon provides real-valued ratings of valence, arousal, and dominance for over 20,000 English words. 
The scores range from 0 to 1 along each of the three dimensions: valence (from maximally unpleasant to extremely pleasant), arousal (from maximally calm, sleepy to maximally active, intense), and dominance (from maximally weak to maximally powerful). 
The annotations were obtained through crowd-sourcing. 

We consider words that appear in the VAD lexicon, and count the percentage of words that have high/low valence, arousal, and dominance scores. (The SOLO query words are excluded from the analysis.) 
For all three dimensions, we consider scores greater than or equal to $0.75$ as high scores, and scores lower than or equal to $0.25$ as low scores. 
Table~\ref{tab:VAD} shows the percentage of words in the different sub-corpora with high/low valence, arousal, and dominance scores. 
Within each row, all the differences are statistically significant (Chi-squared test,  $p < 0.0001$).

We can see again that the solitude tweets have the highest number of strongly positive words (high valence), and the lonely and loneliness tweets have the most strongly negative words (low valence). 
The loneliness corpus has the highest number of negative words, 72\% more than the solitude corpus. 
The lonely and loneliness sub-corpora also have more high-arousal words than the solitude corpus, while the solitude corpus has the highest amount of low-arousal words. 
The solitude tweets tend to describe quiet and relaxing moments, in natural surroundings, with no agenda to follow. 
When lonely, people can feel scared and anxious, showing more arousal. 
Also, loneliness is associated with both momentary and chronic stress, which may explain why \textit{lonely} occurs among higher arousal words \cite{seeman1996social}. 
The solitude corpus has the most high-dominance words, 56\% more than the lonely corpus and 24\% more than the loneliness corpus. 
This is consistent with the conceptual definition of \textit{solitude} as a positive, voluntary state of being alone. 
In contrast, when feeling lonely, people usually perceive the situation as undesirable, they feel scared, depressed, miserable, and powerless.

\begin{table}[t]
{\small
\begin{center}
\begin{tabular}{lrrrr}
\hline
\textbf{Dimension} & \textbf{general} & \textbf{loneliness} & \textbf{lonely} & \textbf{solitude}\\
\hline
Valence\\
$\ \ $ low & 9.3 & \textbf{15.8} & 12.3 & 9.2\\
$\ \ $ high & 29.4 & 30.2 & 30.3 & \textbf{33.7}\\
Arousal\\
$\ \ $ low & 9.1 & 10.9 & 11.5 & \textbf{14.4}\\
$\ \ $ high & 8.3 & \textbf{8.6} & 7.2 & 6.2\\ 
Dominance\\
$\ \ $ low & 4.8 & 8.3 & \textbf{8.5} & 7.1\\
$\ \ $ high & 11.9 & 9.9 & 7.9 & \textbf{12.3}\\
      \hline
\end{tabular}
\caption{The percentage of words with high/low valence, arousal, and dominance scores in the SOLO corpus. `general' stands for `General Tweets Corpus'. The highest numbers in each row are in bold. Within each row, all the differences are statistically significant (Chi-squared test,  $p < 0.0001$).}
\label{tab:VAD}
 \end{center}
 }
\end{table}

\begin{figure}[t]
{\small
\begin{center}
\includegraphics[scale=0.33]{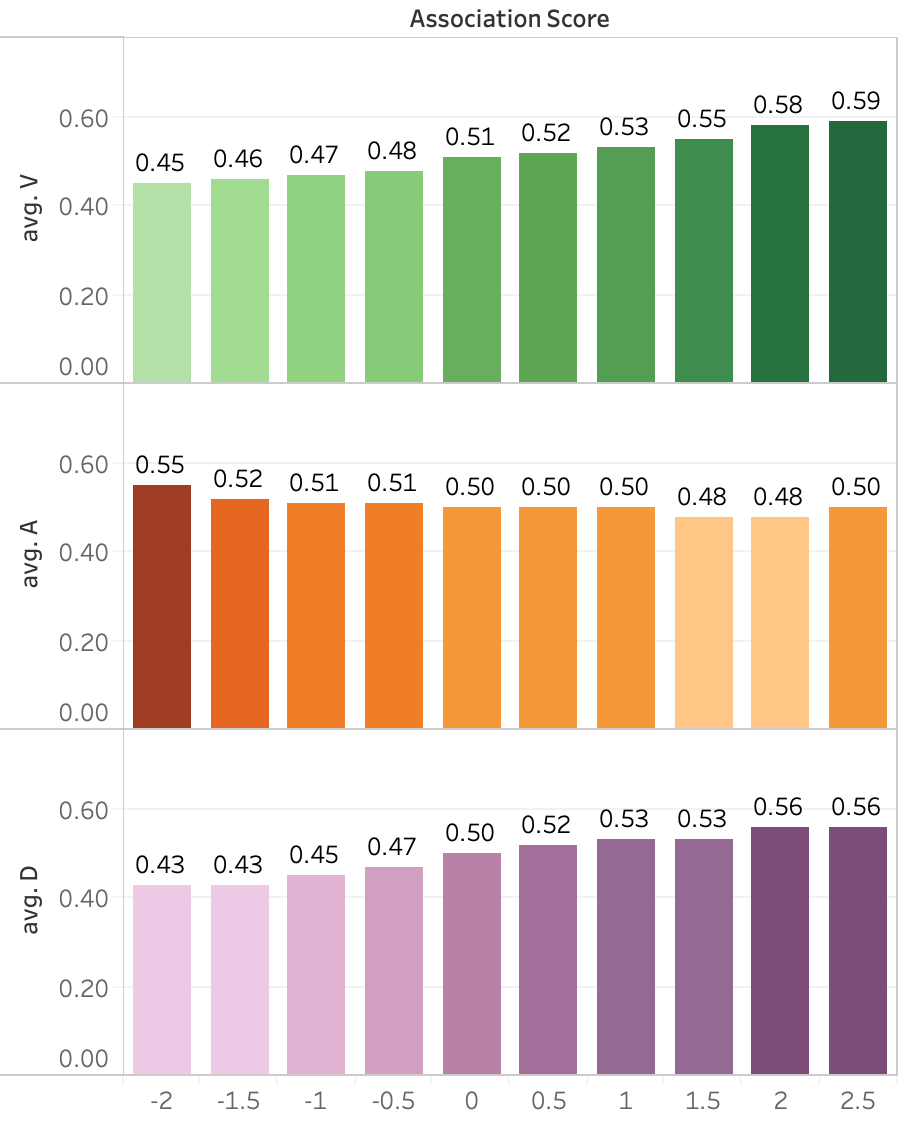} 
\caption{Trends in average valence (V), arousal (A), and dominance (D) scores along the solitude--loneliness dimension of association. 
Positive association scores indicate the word's stronger association with solitude than with loneliness; negative association scores indicate stronger association with loneliness than with solitude.}
\label{VAD-sol-vs-lon}
\end{center}
}
\end{figure}

\vspace{5pt}
\noindent \textbf{VAD Trends Along the Solitude--Loneliness Dimension of Word Association:} We analyze the trends in valence, arousal, and dominance scores along the solitude--loneliness dimension.  
We use the solitude--loneliness association scores for words computed as described in Section~\ref{lang-analysis} 
We order the words by their solitude--loneliness association scores from smallest to largest, bin the scores with a 0.5 step, and average the valence, arousal, and dominance scores for all words that fall in each bin. 
For example, for bin with the score of 1 we average the VAD scores of all the words whose association scores fall in the range [$0.75$, $1.25$). 
The VAD scores for words are taken from the NRC VAD Lexicon. 
Figure~\ref{VAD-sol-vs-lon} shows the trends in the average VAD scores along the solitude--loneliness dimension. 
(Only bins with at least 100 words are shown.) 
Recall that words with positive association scores occur at a higher rate in the solitude sub-corpus and at a lower rate in the lonely and loneliness sub-corpora while the words with negative association scores occur at a higher rate in the lonely and loneliness sub-corpora and at a lower rate in the solitude sub-corpus. 
Along all three dimensions (valence, arousal, and dominance), the trends are very consistent: the more the word is associated with \textit{solitude}, the higher its valence and dominance scores are, and the lower its arousal score is. 
While the range of the average arousal scores is relatively small (from $0.48$ to $0.55$), the differences in the average valence and dominance scores are substantial (from $0.45$ to $0.59$ for valence, and from $0.43$ to $0.56$ for dominance). 
This once again supports the hypothesis that solitude is often viewed as positive, intrinsically motivated state of being alone, and loneliness is viewed as negative, externally imposed state of being alone.  

\section{Demographic Differences in the Language Associated with the State of Being Alone}

In this section, we examine the differences in the language and emotions associated with the state of being alone between genders (male vs. female) and age groups (adolescents vs. adults). 
Researchers have long been interested in exploring differences in language use between  genders in different communication media and sociocultural contexts \cite{park2016women,coates2015women}. 
Here, we continue this line of work and investigate whether men and women tend to use the SOLO concepts, \textit{loneliness}, \textit{lonely}, and \textit{solitude}, in different emotional contexts.
Psychologists are also interested in identifying developmental differences in the perception and experiences with the state of being alone \cite{coplan2019does}. 
Using the large amounts of tweets in the SOLO Corpus and an existing word--age association lexicon, we analyze the tendency of different age groups to describe their experiences of being alone as solitude or loneliness states.

\subsection{Gender Differences in the Language Associated with the State of Being Alone}

To infer the gender of the tweeters, we use the US Social Security Administration database\footnote{https://www.ssa.gov/oact/babynames/limits.html. We acknowledge that users may identify their gender as non-binary, but we did not have the data to explore this. We also acknowledge that US Social Security information is not representative of the names from around the world. 
Thus, the gender analysis is mostly representative of US residents.}. 
From the database, we select first names that occur more than 100 times in total over the years from 1940 until 2017 and that were used for males (females) at least 95\% of the times. 
In total, we found 19,714 female and 10,909 male such names. 
We split the user names of the tweeters by punctuation marks and match the first token against the selected first names. 
If the first token matches one of the female (male) first names, the user is considered female (male). 

\begin{table}[t]
{\small
\begin{center}
\begin{tabular}{lrr}
\hline
\textbf{Corpus} & \textbf{Total tweets} & \textbf{Tweets with}\\
 & &\textbf{inferred gender}\\
 \hline
General Tweets & 21,719,409 & 8,355,543 (38\%)\\ [5pt]    
SOLO Corpus:\\
$\ \ $ loneliness  &   489,264 & 169,305 (35\%)\\  
$\ \ $ lonely & 3,339,166 & 1,131,935 (34\%)\\   
$\ \ $ solitude     &   191,643 & 68,721 (36\%)\\   
      \hline
\end{tabular}
\caption{The total number of tweets with inferred gender of the tweeter.}
\label{tab:num-gendered-tweets}
 \end{center}
 \vspace*{3mm}
} 
\end{table}

\begin{table}[t]
{\small
\begin{center}
\begin{tabular}{lrr}
\hline
\textbf{Corpus} & \multicolumn{2}{c}{{Tweets written by}} \\
& \textbf{Females} & \textbf{Males}\\ 
\hline
General Tweets & 3,730,986 (45\%) & 4,624,557 (55\%)\\   [5pt]    
SOLO Corpus:\\
$\ \ $ loneliness  &   87,228 (52\%) & 82,077 (48\%)\\  
$\ \ $ lonely & 636,388 (56\%) & 495,547 (44\%)\\ 
$\ \ $ solitude     &   33,000 (48\%) & 35,721 (52\%)\\   
      \hline
\end{tabular}
\caption{The number of tweets written by (inferred) female and male users.}
\label{tab:num-F-M-tweets}
 \end{center}
}
\end{table}

\begin{figure}[t]
\begin{center}
\includegraphics[scale=0.45]{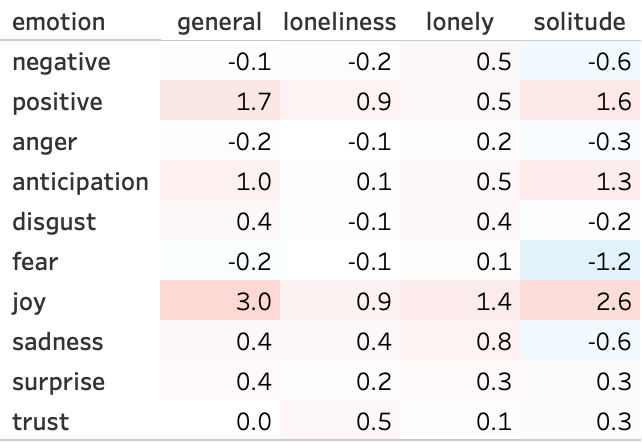} 
\caption{The differences in percentages of words associated with eight basic emotions in tweets written by female and male users. Positive scores (shown in red) indicate that females tend to use more words associated with this emotion than males do. Negative scores (shown in blue) indicate that males tend to use more words associated with this emotion than females do. Darker shades of red/blue highlight differences with larger absolute values. `general' stands for `General Tweets Corpus'.}
\label{fig:VAD-gender}
\end{center}
\end{figure}

Table~\ref{tab:num-gendered-tweets} shows the number of tweets with inferred tweeter gender for each sub-corpora. 
We are able to infer gender of the tweeter in 34\%--38\% of the tweets. 
Table~\ref{tab:num-F-M-tweets} shows the percentage of tweets written by female and male users. 
Notice that in the General Tweets, the majority of the tweets with inferred gender is from male users (55\%). 
Similar percentage of male users is inferred in the solitude sub-corpora (52\%). 
However, in the lonely and loneliness sub-corpora the majority of the inferred users are female (56\% and 52\%, respectively). 
This suggests that women have and/or report their negative experiences of being alone more often than men. 
 
To examine the differences in emotional content of tweets written by different genders, we perform analyses of basic emotions and valence, arousal, and dominance in a similar manner as described in Section~\ref{emotion-analysis} 
The analyses are performed separately on the tweets written by male users and on the tweets written by female users. 
Figure~\ref{fig:VAD-gender} shows the differences in percentages of words associated with eight basic emotions in tweets written by female and male users. 
Observe that in the General Tweets Corpus the differences are minor, most of them are below 1\%. 
The only differences that are 1\% or larger are for the emotions of joy (3\% more in text written by women) and anticipation (1\% more in text written by women) as well as for positive sentiment (1.7\% more in text written by women).
We see similar trends in the solitude sub-corpus: the only differences that are larger than 1\% in absolute values are for the emotions of joy, anticipation, fear, and for positive sentiment. 
In the lonely and loneliness sub-corpora, the differences across genders are even smaller---below 1\% for all, except for the emotion of joy in the lonely sub-corpus (1.4\%). 
The results for valence, arousal, and dominance are also similar
(numbers not shown here).  
Overall, within tweets associated with the state of being alone, the differences in emotional content across the two genders are small.

\subsection{Age Differences in the Language Associated with the State of Being Alone}

\setlength{\tabcolsep}{5pt}

\begin{table}[t]
{\small
\begin{center}
\begin{tabular}{lrrrr}
\hline
\textbf{Corpus} & \multicolumn{4}{c}{{Percentage of words associated}}\\
& \multicolumn{4}{c}{{with an age group:}}\\
 & \textbf{13 to 18} & \textbf{19 to 22} & \textbf{23 to 29} & \textbf{30+}\\
\hline
General Tweets & 31.0  &   5.1 & 10.5 & 55.3\\ [5pt]
SOLO Corpus:\\
$\ \ $ loneliness & 29.8  &   5.4 & 9.8 & 54.4\\
$\ \ $ lonely & 37.5  &   7.4 & 8.8 & 48.2\\
$\ \ $ solitude & 27.4   &  4.6 & 11.0 & 57.2\\
\hline
\end{tabular}
\caption{Percentage of words associated with different age groups. Within each age group (column), all the differences are statistically significant (Chi-squared test,  $p < 0.0001$).}
\label{tab:age}
 \end{center}
 }
\end{table}

\setlength{\tabcolsep}{6pt}

Since we do not have age information for the tweeters in our corpus, we use an available Word--Age Association Lexicon \cite{schwartz2013personality}. 
This lexicon provides association scores and the corresponding p-values for common words and phrases (1-grams, 2-grams, and 3-grams) with four age groups: 13 to 18 years old, 19 to 22 years old, 23 to 29 years old, and 30 and over years old. 
\newcite{schwartz2013personality} collected Facebook messages of 75,000 volunteers, along with the information on their age and gender. 
Then, they calculated the association scores by fitting a linear function between the target variable (word's relative frequency) and the dependent variable (age), adjusted for gender. 
The lexicon includes only those words and phrases that were used by at least 1\% of all subjects.  
From the lexicon, for each age group, we collect single, alpha-numeric tokens that are significantly positively associated with the age group ($p \leq 0.05$). 
Out of 8,093 single, alpha-numeric tokens in the lexicon, 1,921 were significantly positively associated with the 13 to 18 years old group, 845 were significantly positively associated with the 19 to 22 years old group, 1,130 were significantly positively associated with the 23 to 29 years old group, and 3,055 were significantly positively associated with the 30 and over years old group.

Using the Word--Age Association Lexicon, we calculate the percentage of words associated with each age group in each sub-corpus (loneliness, lonely, solitude, and general tweets). 
For this, we divide the number of occurrences of words associated with a particular age group by the total number of occurrences of all the words in the lexicon. 
Table~\ref{tab:age} shows  the results. 
Within each age group, all the differences between the numbers for each sub-corpus (loneliness, lonely, and solitude) and the general tweets are statistically significant (Chi-squared test,  $p < 0.0001$). 
Observe that the lonely sub-corpus has higher percentages of words associated with the two younger groups (as compared with the general tweets) and lower percentage of words associated with the two older groups. 
The differences for the teenage group and the older adults (30+ years old) are particularly large (21\% increase for the teenage group, 13\% decrease for the 30 and over group). 
The solitude sub-corpus shows the opposite pattern with lower percentage of words associated with the two younger groups and higher percentage of words associated with the two older groups.  
The differences between the numbers for the loneliness corpus and the general tweets are relatively small for all four age groups. 
These results suggest that there are more younger people (especially teenagers) among the tweeters that use the word \textit{lonely} when talking about being alone and, therefore, have more negative experiences when alone, than among the tweeters that use the word \textit{solitude} and have more positive attitudes to the state of being alone. 
This finding does not support the psychology literature that proposes that adolescence may be a time when being alone is adaptive and enjoyable \cite{coplan2019does}. 
It is possible, however, that adolescents may use Twitter to vent or share feelings about loneliness more often than other age groups.

\section{Applications}

In this section, we list the potential applications and the directions for future work using the resources created as part of this project: the SOLO Corpus, the lexicons of words associated with the SOLO concept terms, and the list of search terms related to the concept of being alone.

\vspace{10pt}
\noindent \textbf{SOLO Corpus:} The corpus can be used to further study how people understand and experience the state of being alone, and how these vary across situations, individuals, and development. For example, the following research questions can be addressed:
\begin{itemize}
	\item  How do people understand different experiences that could be considered `solitary'? Do people distinguish between different degrees of solitude? For example, is someone more `alone' if they are away from their Smartphone? 
	\item What are the different motivations (intrinsic and extrinsic) for people to spend time alone?
	\item Do people recognize some solitary experiences as being more beneficial or costly than others? What are the different benefits that might arise from being alone (e.g., creativity, relaxation, productivity)?
	\item Can we identify developmental differences in experiences and attitudes towards being alone?
\end{itemize}

\vspace{5pt}
\noindent  \textbf{Words associated with the SOLO concept words:}	Words highly associated with the terms \textit{loneliness}, \textit{lonely}, and \textit{solitude} can be used to identify pieces of text that do not necessarily mention either of these three words, but nevertheless discuss the experiences of being alone. 
This can apply to tweets, but also to other types of text (blogs, emails, novels, etc.). 
For example, texts rich in words highly associated with \textit{lonely} have a high probability of discussing feelings of being lonely and the related issues even if the word \textit{lonely} itself is not mentioned. 

\vspace{8pt}
\noindent \textbf{Search terms:} We have shown that by using the search terms \textit{loneliness}, \textit{lonely}, and \textit{solitude} we can collect voluminous corpora of tweets highly related to the state of being alone. Therefore, this search strategy over the Twitter stream can be used to monitor the positive and negative aspects of being alone and their relation to well-being over the entire population across time, geographical regions, and demographic groups.

\vspace{8pt}

\noindent {\bf Building Other SOLO Corpora:} The approach presented in this paper can also be used to create other more focused corpora pertaining to specific demographics for whom solitude and loneliness are particularly relevant, such as the elderly and teenagers \cite{luhmann2016age,hawkley2015perceived}.

\section{Conclusion}

We presented the \textit{SOLO} (\textit{State of Being Alone}) corpus---a large corpus of tweets associated with the state of being alone. 
SOLO includes over 4 million tweets collected using one of the three terms \textit{loneliness}, \textit{lonely}, and \textit{solitude}. 
Manual examination showed that the corpus contains over 94\% of the tweets related to the concept of being alone. 

We used the SOLO Corpus to examine the language and emotions associated with the state of being alone. 
We found evidence that Twitter users tend to use the word \textit{solitude} to describe more positive and self-imposed states of being alone, and tend to use the words \textit{lonely} and \textit{loneliness} when their experiences are negative and undesirable, which is consistent with conceptual definitions proposed in psychology literature. 
Furthermore, we found that the word \textit{loneliness} tends to be used in more negative contexts than the word \textit{lonely}. 

Over the same period of time, the term \textit{lonely} triggered 17 times more tweets than the term \textit{solitude}. 
There were 12\% more tweets with the word \textit{lonely} written by female users than tweets written by males, even though in the General Tweet Corpus (used as control) there were 10\% more tweets written by male users. 
However, the emotional content in the SOLO tweets written by male and female users was strikingly similar. 
We also found more words associated with the adolescent age group (especially, teenagers) and less words associated with the adult age group in the lonely corpus as compared to the solitude corpus, which suggests a higher vulnerability of teenagers to the negative experiences of feeling lonely.

We make SOLO and other resources created in this project freely available to encourage further research on health, economical, and other issues related to people's experiences of being alone and how these issues affect the population's well-being.

The current study focused on English-language social media, in particular tweets. 
In future work, texts from other genres, such as blogs, news, poetry, and fiction, can be analyzed in a similar manner. 
While this study examined the percentage of basic emotion words, one can also use lexica such as the NRC Emotion Intensity Lexicon \cite{LREC18-AIL} to examine the use of high and low intensity emotion words in expressions of solitude.\footnote{http://saifmohammad.com/WebPages/AffectIntensity.htm}
By comparing sources from different time periods, we can track how people's perception of solitude and loneliness change over time. 
Furthermore, parallel studies in other languages can shed light on cultural differences in people's attitudes towards and experiences with the state of being alone.

Finally, we are exploring the creation of corpora similar to SOLO with a focus on text generated by specific demographics such as teenagers, elderly, as well as, those coping with disabilities, stress, or other mental and physical conditions. We believe that a better understanding of people's attitudes towards solitude and loneliness will help identify new ways to improve their well-being. 

\vspace{10pt}
\section*{Acknowledgments}
We thank Samuel Larkin for help in collecting tweets.

\vspace{10pt}
\section{Bibliographical References}
\label{main:ref}

\bibliographystyle{lrec}
\bibliography{solitude}

\end{document}